\documentclass[10pt,twocolumn,letterpaper]{article}
\newif\ifarXiv
\arXivtrue 

\usepackage{iccv}
\usepackage{times}
\usepackage{epsfig}
\usepackage{graphicx}
\usepackage{amsmath}
\usepackage{amssymb}

\usepackage{booktabs}
\usepackage{multirow}
\usepackage{makecell}

\ifarXiv
\iccvfinalcopy
\else
\fi

\ificcvfinal
\newcommand{\yb}[1]{}

\newcommand{\myparagraph}[1]{\vspace{-10pt}\paragraph{#1}}
\else
\newcommand{\yb}[1]{\textcolor{magenta}{[Yuval: #1]}}

\newcommand{\myparagraph}[1]{\vspace{-10pt}\paragraph{#1}}
\fi
\usepackage[pagebackref=true,breaklinks=true,letterpaper=true,colorlinks,bookmarks=false]{hyperref}

\ifarXiv
\usepackage[title]{appendix}
\else
\fi

\def\iccvPaperID{7137} 

\ificcvfinal\fi

\begin{document}

\title{Neural Volume Super-Resolution}

\author{Yuval Bahat\textsuperscript{1} \qquad Yuxuan Zhang\textsuperscript{1} \qquad Hendrik Sommerhoff\textsuperscript{2} \qquad Andreas Kolb\textsuperscript{2} \qquad Felix Heide\textsuperscript{1}\vspace{5pt}\\
\textsuperscript{1}Princeton University \quad \textsuperscript{2}University of Siegen\\
}

\maketitle
\ifarXiv\else
\thispagestyle{empty}
\fi
\ificcvfinal\thispagestyle{empty}\fi

\begin{abstract}
Neural volumetric representations have become a widely adopted model for radiance fields in 3D scenes. These representations are fully implicit or hybrid function approximators of the instantaneous volumetric radiance in a scene, which are typically learned from multi-view captures of the scene. We investigate the new task of neural volume super-resolution -- rendering high-resolution views corresponding to a scene captured at low resolution. To this end, we propose a neural super-resolution network that operates directly on the volumetric representation of the scene. This approach allows us to exploit an advantage of operating in the volumetric domain, namely the ability to guarantee consistent super-resolution across different viewing directions. To realize our method, we devise a novel 3D representation that hinges on multiple 2D feature planes. This allows us to super-resolve the 3D scene representation by applying 2D convolutional networks on the 2D feature planes. We validate the proposed method by super-resolving multi-view consistent views on a diverse set of unseen 3D scenes, confirming qualitative and quantitatively favorable quality over existing approaches.
\end{abstract}
\section{Introduction}
\label{sec:intro}
Reconstructing latent high-quality images from images captured under non-ideal imaging conditions is a broadly studied research field. A large body of existing work explores methods for removing image noise \cite{buades2005non_local_denoising,kawar2021stochastic_denoising}, sharpening blurry images \cite{tao2018scale_recurrent_deblur,bahat2017deblurring_reblurring}, and increasing image resolution \cite{michaeli2013blind_recurrence,ledig2017srgan}. This direction includes recent works that are based on deep learning~\cite{ledig2017srgan,tao2018scale_recurrent_deblur}, mapping degraded images to their latent non-degraded counterparts, and have led to a significant leap in performance. At the heart of these methods typically lies the application of a series of 2D convolution operations to the degraded image \cite{lim2017edsr,ledig2017srgan}, which eventually yields the reconstructed image at a higher resolution.

Concurrently with research on image reconstruction, a rapidly growing body of work explores neural rendering, allowing not only to reconstruct a single image but unseen novel views from a set of observed images. In their work on Neural Radiance Fields (NeRF) \cite{mildenhall2020nerf} in 2020, Mildenhall \etal showed that state-of-the-art rendering quality can be achieved even with a lean Multilayer Perceptron (MLP) architecture as a coordinate-based representation for a 5D radiance field. Many works have since proposed improved and more efficient ways to represent 3D scenes based on a given finite set of 2D images, including methods that tackle modified lighting conditions \cite{srinivasan2021nerf_relight} or support dynamic scenes \cite{ost2021scene_graphs}. These existing methods directly supervise the scene representation using the observed images.

\begin{figure}[t]
\centering
\includegraphics[width=\linewidth]{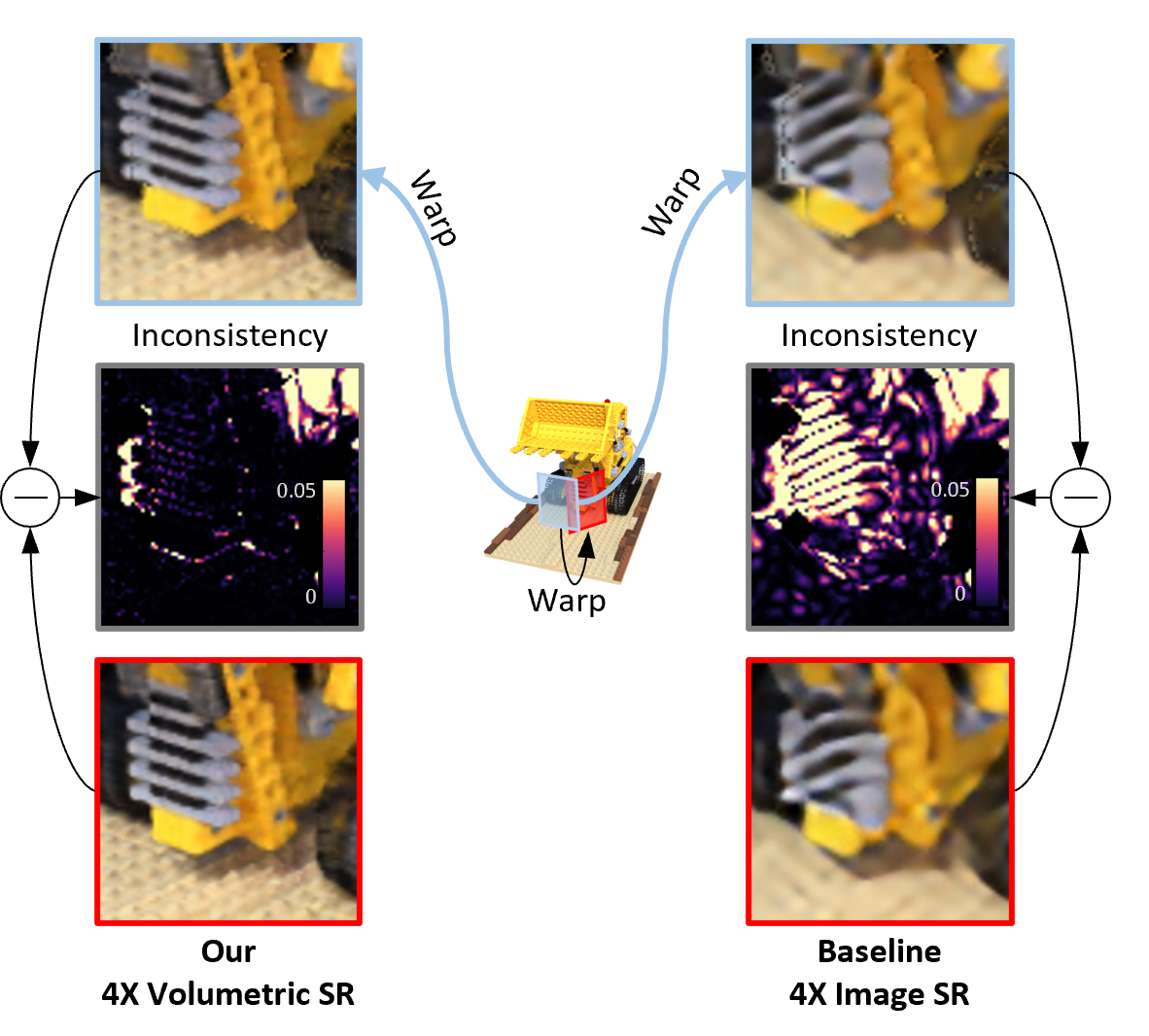}
\caption{\label{fig:teaser}\textbf{Geometrically consistent super-resolution.}
Our method (bottom left) renders HR views of a scene captured in LR by super-resolving its volumetric representation. This guarantees the geometric consistency of the rendered scene across viewing directions, and prevents inconsistent distortions such as the slanted bulldozer grill in the output of the EDSR~\cite{lim2017edsr} image SR method (bottom right).
Here we visualize the across-views inconsistency by rendering two adjacent high-resolution scene views, and then warping one of them (top) onto the perspective of the other (bottom). We then compute error maps between each rendered pair (middle row), which further highlights our enhanced consistency.
}
\end{figure}

In this work, we explore an alternative direction 
at the intersection of image reconstruction and neural rendering, and propose the new task of neural volume super-resolution - learning to super-resolve neural volumes.
Specifically, given a finite set of low-resolution images corresponding to a 3D scene, we aim to render high-resolution images of the scene corresponding to arbitrary possible viewing directions, including novel unseen ones. Rather than operating on the rendered (2D) images and upsampling these images after generation, we propose a novel approach to tackle this task, by operating directly on the volumetric (3D) scene representation. An advantage of this approach, illustrated in Fig.~\ref{fig:teaser}, is that it guarantees by design a \emph{geometrically consistent} image reconstruction across all viewing directions, in contrast to operating in the 2D domain and separately reconstructing each rendered image.

As a practical and efficient approach for enabling upsampling neural volumes, we propose a new 2D plane-based scene representation that allows us to employ 2D convolutional neural networks. Specifically, we represent a 3D scene by implicitly associating each point in the 3D volume with a density value and a (view-direction dependent) RGB value following existing radiance fields works, while investigating an alternative way to infer these values. Chan \etal \cite{chan2022eg3d} accumulated features for each 3D point by projecting its location onto three \emph{2D} feature planes, for representing human faces. We instead propose a plane representation that allows representing general scenes, and introduce a fourth feature plane to enable view direction dependency. Given a desired viewing direction, we can reconstruct each image pixel by marching along its corresponding ray from the camera, taking into account the cumulative effect of the density and RGB values predicted by our model for each point along the ray. 

Rendering a \emph{super-resolved} image from the proposed scene model, that was learned from a set of \emph{low-resolution} images, requires a single modification to our pipeline, while the rest remains unchanged: Before projecting each point in the volume onto the 2D planes to infer its density and RGB values, we super-resolve the three 2D planes using a learned super-resolution module that operates in feature space. This approach allows us to take advantage of existing methods explored for 2D super-resolution over the years. 

We train our \emph{Neural Volume Super-Resolution} (NVSR) framework end-to-end using low-resolution and matching high-resolution image pairs corresponding to synthetic scenes from freely available online datasets.
We then apply the trained SR model to the 3D representations of unseen scenes for which only low-resolution images are available. We validate that unlike super-resolving images post-rendering, the resulting super-resolved rendered images are indeed consistent across different views. The proposed method outperforms such conventional post-rendering SR methods both qualitatively and quantitatively, exceeding state-of-the-art image SR methods by over $1$~dB in PSNR when trained on the same dataset.

Specifically, we make the following contributions:
\begin{itemize}
    \item We propose a method for super-resolving volumetric neural scene representations, which allows generating geometrically consistent, high-resolution novel views corresponding to a scene captured in low resolution. 
    \item We introduce a super-resolution approach that operates directly in feature space. To this end, we develop a novel feature-plane representation model that allows for rendering viewing-direction dependent views of general 3D scenes. 
    \item We assess our method for super-resolving unseen scenes both qualitatively and quantitatively, validating that it produces sharp, multi-view consistent super-resolved images.
\end{itemize}
We will release all code and models needed to reproduce the results presented in this work. 
\section{Related Work}\label{sec:related_work}
This work introduces a method for reconstructing images of a 3D scene by operating directly on its volumetric neural representation. Next, we provide background on the two most related fields, image super-resolution and neural 3D scene representation.

\myparagraph{Image Super Resolution}
Image super resolution (SR) belongs to a wide category of ill-posed image reconstruction tasks such as image denoising~\cite{buades2005non_local_denoising,kawar2021stochastic_denoising}, dehazing~\cite{li2018benchmarking_dehazing,bahat2016recurrence_dehazing}, deblurring~\cite{tao2018scale_recurrent_deblur,bahat2017deblurring_reblurring}, inpainting~\cite{bertalmio2000image_inpainting,jam2021inpainting_review} and more. Image SR aims to compensate for information lost during capturing as a result of the lossy image acquisition process, due to a resolution loss in the optical system and finite sampling by the sensor. Existing image SR methods produce a reconstructed image of the captured scene as if captured in ideal settings.
To recover the fine details lost due to the finite capturing resolution, some SR works propose to collect additional information from multiple captures of the same scene~\cite{farsiu2004fast_multiframe,wronski2019handheld_burst}, which typically allows increasing the resolution by only up to a factor of two. To allow higher SR factors, classical methods have proposed to employ various image priors that exploit the unique characteristics of natural images, \eg 
their internal self-similarity~\cite{glasner2009recurrence,michaeli2013blind_recurrence}, heavy-tailed gradient magnitude distribution~\cite{sun2008gradient_prior} and distinctive image edge statistics~\cite{tai2010edge_prior}. 

\myparagraph{Learning Image Super Resolution}
With the arrival of deep learning, researchers proposed to learn these image characteristics and train Convolutional Neural Networks (CNNs) on large image datasets~\cite{dong2014first_cnn_sr,lim2017edsr,liang2021swinir}, leading to unprecedented performance in terms of minimizing reconstruction error (\ie increasing PSNR). Following the introduction of Generative Adversarial Networks (GANs)~\cite{goodfellow2020gan}, some methods also began targeting alternative objectives, such as perceptual reconstruction quality~\cite{ledig2017srgan,wang2018esrgan} or exploring the diverse space of SR solutions~\cite{bahat2020explorable,lugmayr2020srflow}.
A further line of work on image super resolution takes relevant reference images as additional user input~\cite{zheng2018reference,dong2021rrsgan_reference} to better adapt to each specific scene. However, common to all of these methods is that they aim to super-resolve a \emph{single} image at a time. To the best of our knowledge, the proposed is the first work aiming to \emph{consistently} super-resolve images corresponding to the \emph{infinite} possible views of a given 3D scene, by operating directly on its neural representation. While methods for spatial Video SR~\cite{kappeler2016video_sr,liu2017robust_video_sr,geng2022video_sr} also attempt to
simultaneously handle sequences of inter-related frames, these methods are limited to positions along the original camera trajectory  and cannot render novel views. Moreover, they cannot inherently guarantee consistent reconstruction across frames, unlike our approach.

\myparagraph{Neural Scene Representations.}
An increasingly large body of work learns representations of 3D scenes, aiming to allow synthesizing novel scene views from arbitrary directions, as well as editing the scene \eg to allow modifying the perceived illumination settings or object properties.
Early works~\cite{agarwal2011building_rome,schonberger2016pixelwise} tried to learn an explicit 3D scene model by fitting it to a given set of images capturing the scene from different view directions. 
The recent introduction of Neural Radiance Fields~\cite{mildenhall2020nerf} (NeRF) then brought upon a significant leap in reconstructed image quality, rendering photo-realistic views which include fine, high-frequency details. They rely on a lean coordinate-based Multilayer Perceptron (MLP) to implicitly represent the 5D radiance field, made possible mainly thanks to artificially introducing high frequencies into the model input, in the form of sinusoidal positional encoding~\cite{tancik2020pos_encoding}.

In their follow up Mip-NeRF~\cite{Barron_2021mipnerf,barron2022mipnerf360} papers, Barron \etal focused on handling diverse scene capture resolutions. By modeling sampled rays as conical frustums they were able to reduce blur and aliasing in rendered outputs. 
In contrast to their work, ours allows rendering novel high-resolution views corresponding to a scene which was originally captured in low-resolution, by \emph{reconstructing} the high-frequency content that was lost during scene acquisition.

Other methods proposed to use alternative implicit representation models based on, \eg, Voxels~\cite{liu2020sparse_voxels} or spherical harmonics~\cite{Fridovich-Keil_2022plenoxels}. Chan \etal \cite{chan2022eg3d} proposed to generate geometrically consistent human faces by representing the volume using three 2D feature planes coupled with a simple MLP decoder. We also employ 2D feature planes to support general 3D scenes and view-direction dependency and take advantage of the planar representation to perform our volumetric SR in 2D, while leveraging the existing rich knowledge base which already exists for 2D SR (see Sec.~\ref{sec:planar_representation}).

Many recent methods focused on enabling NeRF to learn from imperfect image sets, \eg using partly occluded~\cite{martin2021nerf_in_the_wild} or very few scene captures~\cite{jain2021few_shot_nerf}. In contrast, our work aims to endow the learned representation model with the ability to reconstruct finer details than the ones captured in the given image sets, by learning volumetric scene priors, \ie, a volumetric extension of (natural) image priors. 

\myparagraph{Neural Radiance Fields for Image Reconstruction.}
In a separate line of work, some methods harness the implicit 3D representation of NeRF as a tool for image denoising~\cite{pearl2022nerf_denoising} or high dynamic range (HDR) reconstruction~\cite{mildenhall2022nerf_in_the_dark,huang2022hdr_nerf}. These methods take as input a burst of noisy or low dynamic range scene captures, respectively, and distill the multi-frame signal while relying on the geometrical consistency inherent to the 3D representation.
In contrast, our approach allows using a set of degraded (low-resolution) scene captures for synthesizing consistent novel scene views, containing fine details that are \emph{unavailable} in the input images, by relying on \emph{externally learned} volumetric priors
(see Sec.~\ref{sec:volumetric_sr}).
\section{Super-Resolving Neural Volumes}
\begin{figure*}[t!]
\centering
\includegraphics[width=\textwidth]{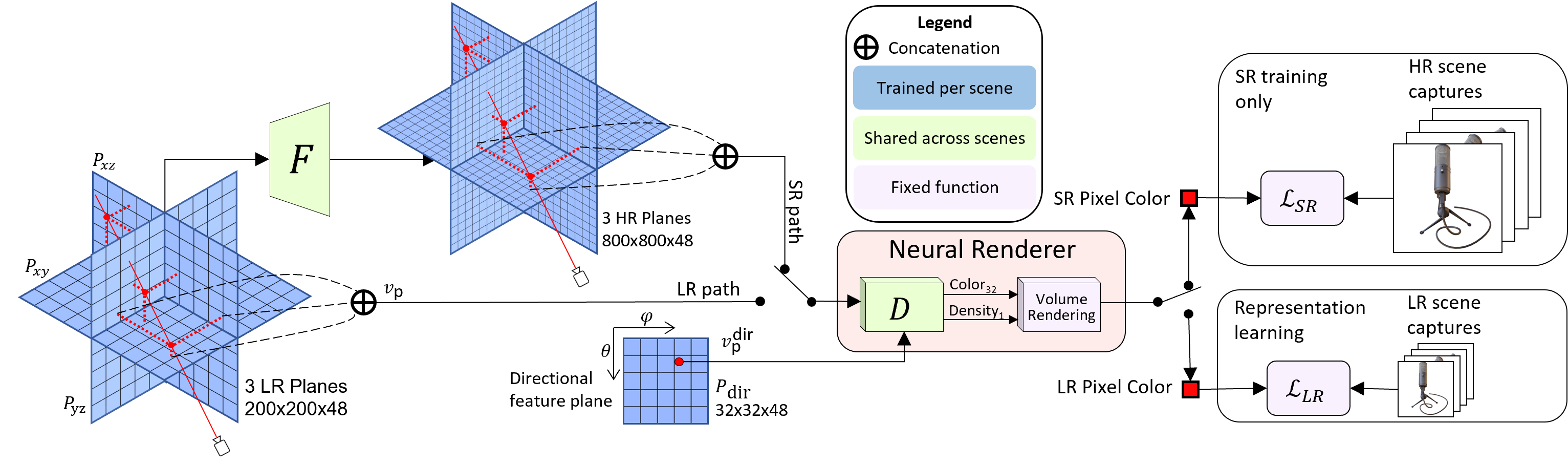}
\caption{\label{fig:overview}\textbf{Overview of our Volumetric SR framework.}
We use three LR positional feature planes (left) and one view-direction feature plane (bottom) to represent each 3D scene based on its captured LR images. Image pixels are rendered by projecting points along the corresponding camera ray onto the four planes and extracting the corresponding feature vectors $v_p$. These are then processed by a decoder MLP $D$ (shared across all scenes) to yield volumetric density and radiance values, which translate into pixel values through volume rendering. To render HR scene views, we first super-resolve the positional feature planes using $F$, and then extract $v_p$ from the super-resolved planes (top). 
We use pairs of GT LR-HR image captures from a set of training scenes to train our framework end to end.
}
\end{figure*}

We introduce a method for super-resolving the representation of a 3D scene  $s$ given ${\cal I}_{\text{lr}}^s$, a set of low-resolution (LR) captures of the scene and their corresponding relative camera poses. To this end, we devise two principle modules, namely a scene representation model $G$ and a super-resolution model $F$.
We first use the images in ${\cal I}_{\text{lr}}^s$ to learn the parameters $\theta_G^s$ of representation model $G$,
which comprises a decoder model $D$ and learned feature planes ${\cal P}_s$
corresponding to each scene $s$. Once trained, one can use $G$ to render novel low-resolution views of the scene for new (unseen) camera poses:
\begin{equation}\label{eq:rendering_lr}
    {\hat i}_{\text{lr}}(r)=G(\theta_G^s,r).
\end{equation}
We use $r$ here to denote a pixel in the rendered image and its corresponding camera ray interchangeably, and denote by ${\hat i}_{\text{lr}}$ the rendered images, as they too lack the high-resolution fine details which are missing from the images in ${\cal I}_{\text{lr}}^s$.
To render high-resolution (HR) views containing fine details and textures, we propose to apply model $F$ on the \emph{parameters} of model $G$ which constitute the volumetric representation of $s$, that is
\begin{equation}\label{eq:rendering_sr}
    {\hat i}_{\text{sr}}(r)=G(F(\theta_F,\theta_G^s),r).
\end{equation}
Here ${\hat i}_{\text{sr}}$ is the rendered super-resolved image and $\theta_F$ are the parameters of model $F$, which are learned independently of any one specific scene.

An overview of our approach is illustrated in Fig.~\ref{fig:overview}. We next elaborate on each of its two modules; we first describe our novel plane-based neural 3D representation model $G$, and then explain how we use this representation to perform volumetric SR using $F$.

\begin{figure}[!b]
\centering
\includegraphics[width=\columnwidth]{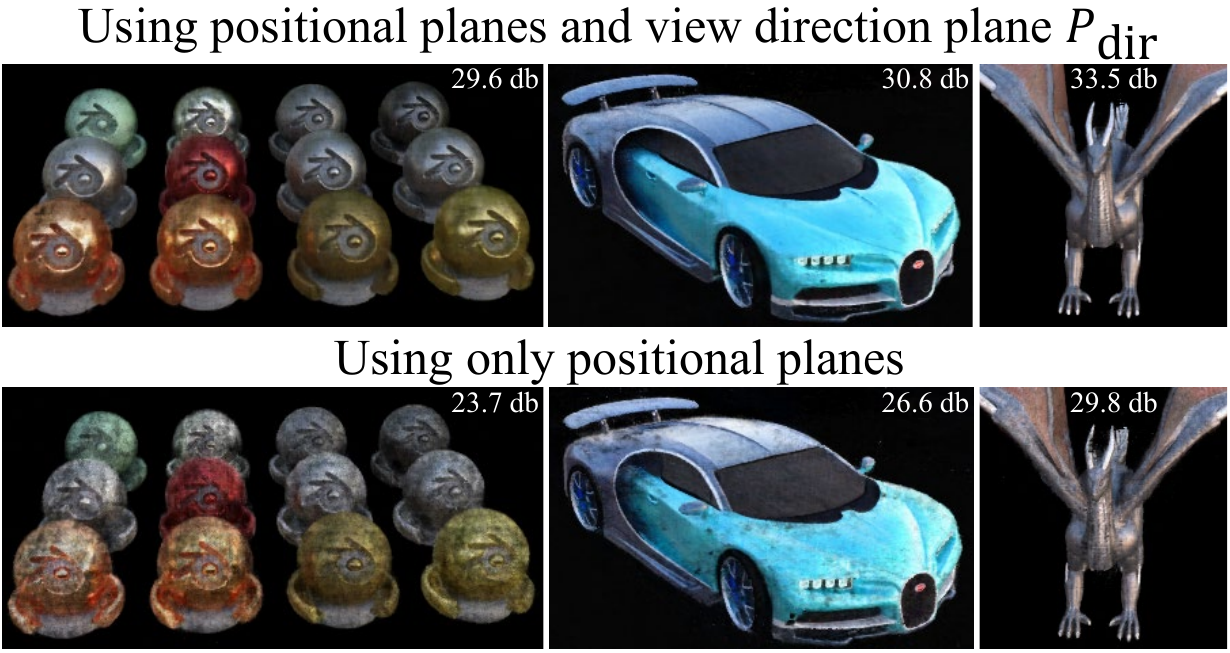}
\caption{\label{fig:viewdir_effect}\textbf{View-direction rendering dependency.}  Utilizing a 4\textsuperscript{th} feature plane $P_\text{dir}$ (top row) corresponding to viewing direction significantly improves reconstruction performance, as it accounts for directional appearance changes stemming from non-uniform lighting conditions, especially in cases of shiny, reflective objects. 
}
\end{figure}

\subsection{Quadri-Plane Radiance Fields for 3D Scenes}\label{sec:planar_representation}
To take advantage of advances in (2D) image SR, and to allow computationally efficient volumetric SR, we propose to represent a scene $s$ with a novel \emph{quadri-plane} model comprised of four multi-channel 2D feature planes of size $N\times N\times C$, coupled with a small decoder neural network.

Similar to existing neural rendering methods \cite{mildenhall2020nerf}, we view the volume as a pair of 3D fields representing scene density $\sigma$ and corresponding emitted radiance (in RGB).
We then diverge from the typical scheme and propose a feature plane-based representation. Chan \etal~\cite{chan2022eg3d} recently proposed a scene representation using feature planes for generating human faces. Specifically, they employ explicit feature vectors arranged in three orthogonal axis-aligned feature planes $\{P_{xy},P_{xz},P_{yz}\}$.
Then each 3D position $p\in\mathbb{R}^3$ is queried by projecting it onto each of the three feature planes and retrieving the per-plane feature vector corresponding to the point via bilinear interpolation, then aggregating them into a point feature representation vector $v_p$. This vector is fed into a light-weight \emph{decoder} MLP $D$ to yield the density and RGB values corresponding to $p$, which are then used to render an image of the scene using neural volume rendering~\cite{mildenhall2020nerf}.
In this work, we propose several modifications to this approach which are critical for the proposed method to function. We describe these modifications in the following.
\myparagraph{View-direction Dependent Radiance.}
We introduce a 4\textsuperscript{th} feature plane $P_\text{dir}$ to accommodate effects like non-uniform scene illumination, as we show in Fig.~\ref{fig:viewdir_effect}. The two axes of this plane reflect the 2D space of possible viewing directions for each point $p$.
We then infer a complementary feature vector $v_p^\text{dir}$ using bilinear interpolation over $P_\text{dir}$ at coordinates $(\theta_p,\phi_p)$, corresponding to the viewing azimuth and elevation of point $p$, respectively.
Features from this plane only affect the radiance and not the density output of $D$, as unlike radiance, volume density does not vary with viewing direction in the real world~\cite{mildenhall2020nerf}.

\myparagraph{Representing \emph{General} Scenes.} To allow representing general scenes with arbitrary geometric structures beyond the human faces demonstrated in~\cite{chan2022eg3d}, we follow the practice from~\cite{mildenhall2020nerf} and use a pair of coarse and fine decoders 
$D_c$ and $D_f$,
respectively,  that are \emph{shared} across all scenes;  we first apply the coarse decoder to a set of stratified sample points $p$ along rays corresponding to each pixel. We then use the resulting values to bias a subsequent sampling of points along each ray towards more relevant parts of the volume, and process those using the fine decoder.

Given a set of training scenes ${\cal S}_t$ and corresponding sets of captured images, we learn a set of four planes per scene ${\cal P}_s=\{P_{xy},P_{xz},P_{yz},P_\text{dir}\},~s\in {\cal S}_t$ jointly with the decoder pair parameters $\theta_D$ (shared across all scenes), which together constitute the representation model parameters $\theta_G^s=\theta_D\cup{\cal P}_s$. 
To this end, we minimize the following image rendering penalty:
\begin{equation}\label{eq:lr_loss}
    {\cal L}_\text{LR}=\hspace{-10pt}\sum_{s\in {\cal S}_t,r\in {\cal I}_{\text{lr}}^s}\hspace{-8pt}
    \|G_c(\theta_G^s,r)-I_\text{lr}(r)\|^2_2+\|G_f(\theta_G^s,r)-I_\text{lr}(r)\|^2_2. 
\end{equation}
Here $I_\text{lr}(r)$ is the ground-truth RGB pixel value corresponding to ray $r$ sampled from the training image $I_{\text{lr}}\in {\cal I}_{\text{lr}}^s$,
and 
$G_c(r)$ and $G_f(r)$ 
are the corresponding rendered values obtained through volume rendering over the outputs of the coarse and fine decoders $D_c$ and $D_f$, respectively.

\begin{figure}[b]
\centering
\includegraphics[width=\columnwidth]{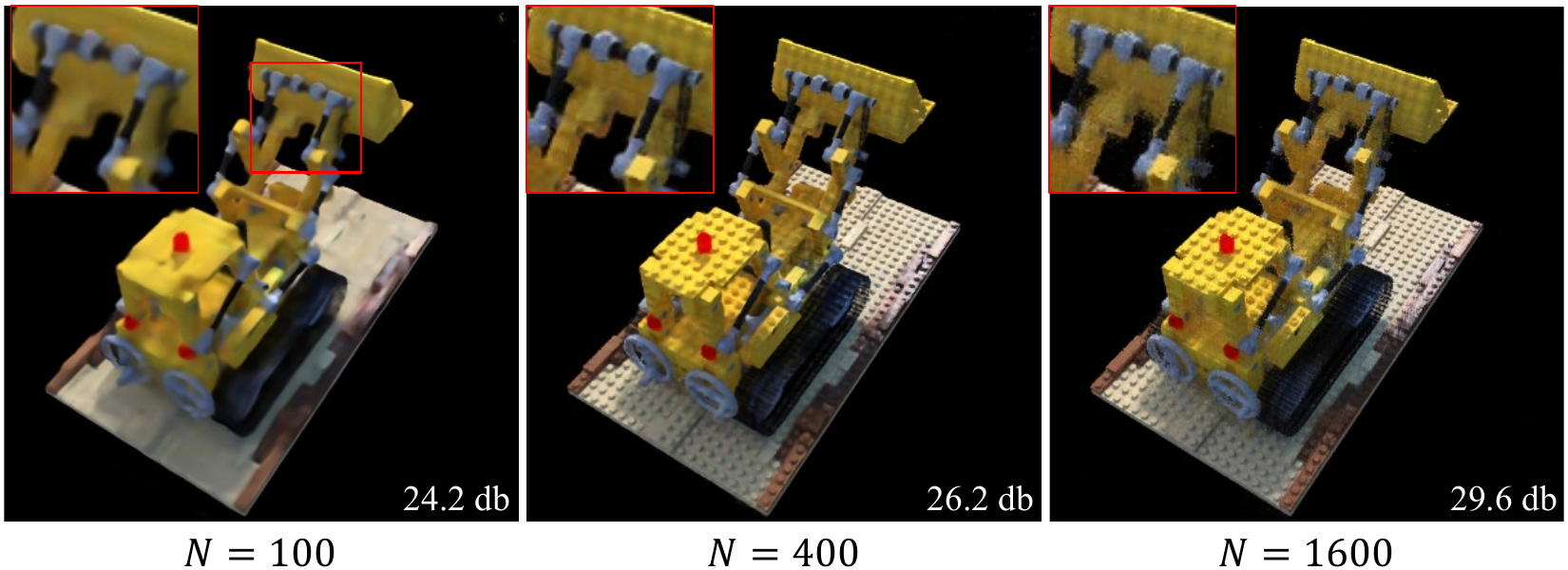}
\caption{\label{fig:plane_res}\textbf{Effect of feature planes resolution.} Novel view reconstructions using different positional plane resolutions $N$, all learned from the same set of $400\times400$ pixel training images. Feature planes that are too small cannot represent fine details (left), while larger planes yield sharp, detailed rendered images (middle). However, at some point increasing the feature plane size leads to visual artifacts (right), as some regions in the oversized feature plane cannot be learned due to the limited resolution of the training images.
}
\end{figure}

We tailor the resolution $N$ of the learned planes to the resolution of the images in ${\cal I}_\text{lr}^s$:
As we show in Fig.~\ref{fig:plane_res}, using planes that are too small prohibits the representation of high visual frequencies, similar to using NeRF~\cite{mildenhall2020nerf} without positional encoding. In contrast, using too large planes results in some plane regions having too little relevant information in the captured scene images ${\cal I}_{\text{lr}}^s$, which is manifested as visual artifacts when rendering novel scene views. We will revisit this trade-off in plane resolution later in the context of super-resolving the scene.

Note also that we choose a smaller resolution value $N_\text{dir}<N$ for the view-direction feature plane $P_\text{dir}$,
since the image set contains less information relevant to this plane:
Each image pixel carries information relevant to all points along its corresponding ray $r$ which, in turn, makes it relevant to many different locations on the positional feature planes $\{P_{xy},P_{xz},P_{yz}\}$. In contrast, all points along a ray correspond to the same viewing direction, meaning that the information in each pixel is relevant to a single point on $P_\text{dir}$.

\subsection{Super-resolving Implicit Representations}\label{sec:volumetric_sr}
To produce HR views of a scene given only a set of its LR captures, one could take a classic super-resolution approach and apply any one of many existing image SR methods on the rendered LR scene views. However, since such methods have no notion of the underlying three-dimensional scene, novel views contain inconsistent super-resolved details in the synthesized views (see Fig.~\ref{fig:teaser}).
Our approach addresses this issue and allows producing super-resolved, \emph{geometrically consistent} scene views. 

As we discuss in Sec.~\ref{sec:planar_representation}, employing larger plane resolution $N$ allows us to represent and render higher visual spatial frequencies, corresponding to small details and fine textures. Therefore, substituting the set of planes ${\cal P}_s$ learned from the LR scene captures ${\cal I}_{\text{lr}}^s$ with a set of higher-resolution planes corresponding to the same scene $s$ can allow for rendering the desired HR views. 

We propose to approximate sets of HR planes by learning plane super-resolution model $F$, which we apply on the learned LR planes (Eq.~\ref{eq:rendering_sr}) to obtain $\alpha$ times super-resolved versions of them, with $\alpha$ being the desired SR factor.
We use a single model to super-resolve all positional\footnote{We use the viewing direction plane as is without super-resolving it.
}
feature planes $P\in{\cal P}_s\setminus\{P_\text{dir}\}$, while taking advantage of our 2D planes-based representation model by employing an existing image SR network architecture for $F$. Specifically, we borrow the residual network architecture used for the successful 2D super-resolution method EDSR~\cite{lim2017edsr}. To learn the mapping from sets of LR planes to their HR counterparts, we train $F$ on a training set of scenes for which ground truth (GT) HR captures are available, by substituting $G_f(r)$ in Eq.~\ref{eq:lr_loss} with the output from Eq.~\ref{eq:rendering_sr} and using HR scene captures $I_\text{hr}\in {\cal I}_{\text{hr}}^s$, resulting in a second loss term:
\begin{equation}\label{eq:hr_loss}
    {\cal L}_\text{HR}=\sum_{s\in {\cal S}_t,r\in {\cal I}_{\text{hr}}^s}\|G_f(F(\theta_F,\theta_G^s),r)-I_\text{hr}(r)\|^2_2.
\end{equation}

\myparagraph{Generalizing to Unseen Scenes.}
When we trained our SR model $F$ through minimizing ${\cal L}_\text{HR}$ over training scenes in ${\cal S}_t$ we observed a severe overfitting problem, which we attribute to the following subtle point: While the input to decoder $D$ corresponding to 3D point $p$ comprises \emph{all three} feature vectors $v_p$, model $F$ acts by 
super-resolving each feature plane \emph{independently}. 
It is trained by minimizing the loss over training scenes in ${\cal S}_t$ and their corresponding $v_p$ triplet combinations. This, however, results in poor performance when tested on unseen scenes \mbox{$s \in {\cal S}_e=\{s|s\notin{\cal S}_t\}$}, consisting of different combinations of $v_p$ which were often not seen during training.

To circumvent this problem we additionally train $F$ on $v_p$ triplets corresponding to test scenes. Since GT HR captures are unavailable for these scenes, we instead exploit their available LR captures, by introducing a downsampling inconsistency loss term ${\cal L}_\text{incon}$. This term penalizes for the difference between the available GT LR views and their corresponding downsampled HR views rendered by our model:
\begin{equation}\label{eq:inconsistency_loss}
    {\cal L}_\text{incon}=\hspace{-10pt}\sum_{s\in {\cal S}_e,r\in {\cal I}_{\text{lr}}^s}\|\big[G_f(F(\theta_F,\theta_G^s),r)\big]\downarrow_\alpha-I_\text{lr}(r)\|^2_2,
\end{equation}
Where $\big[\cdot\big]\downarrow_\alpha$ corresponds to image downsampling by a factor of $\alpha$ and ${\cal S}_e$ denotes the set of test scenes.

We train our framework \emph{end-to-end} and learn the parameters of $G$ (including fine and coarse decoder parameters $\theta_D$ and LR scene planes ${\cal P}_s$ corresponding to each training scene $s$) together with training our SR model $F$. 
To this end, we alternate between the following three optimization steps: (i) Minimizing ${\cal L}_\text{LR}$ \eqref{eq:lr_loss} over LR training scene images $\{{\cal I}_{\text{lr}}^s\}_{s\in {\cal S}_t}$, (ii) minimizing ${\cal L}_\text{HR}$ \eqref{eq:hr_loss} over HR training scene images $\{{\cal I}_{\text{hr}}^s\}_{s\in {\cal S}_t}$, or (iii) minimizing ${\cal L}_\text{incon}$ \eqref{eq:inconsistency_loss} over LR test scene images $\{{\cal I}_{\text{lr}}^s\}_{s\in {\cal S}_e}$. At each training step we randomly choose one of the three at probability ratios of $f_\text{LR}:f_\text{HR}:f_\text{incon}$, respectively.

\subsection{Implementation Details}\label{sec:implementation}
In both of our decoder models $D_c$ and $D_f$, we calculate the density value $\sigma$ corresponding to each 3D point $p$ using fully-connected, $128$ channel, $4$ layer MLPs, which take as input the average of the positional feature vectors $v_p$. For computing the corresponding radiance values $\{r,g,b\}$ we feed the concatenation of all $4$ (including the view-direction) feature vectors $v_p$ into MLPs with identical characteristics.
We use $C=48$ channels and a spatial resolution of $N=200$ for our low-resolution positional feature planes learned from scenes captured at $100$ pixels,
and set the view-direction feature plane $P_\text{dir}$ resolution to $N_\text{dir}=32$.
For our SR model $F$ we use the architecture from EDSR~\cite{lim2017edsr} with $32$ residual blocks and $256$ channels.

We train our entire model end-to-end using Adam optimizers with learning rates of $0.0005$ for the decoder and feature planes and $0.00005$ for $F$. At each step, we sample 4096 rays $r$ from an image $I$ corresponding to a scene $s$, all chosen randomly. 
We use an NVIDIA A100 GPU, and first train our framework to convergence by minimizing ${\cal L}_\text{LR}$ and ${\cal L}_\text{HR}$ over training scenes $s\in {\cal S}_t$ \mbox{($f_\text{LR}=f_\text{HR}=1,~f_\text{incon}=0$)}. Given a test scene $s\in{\cal S}_e$, we resume training by alternating between the three optimization steps at a ratio of \mbox{$f_\text{LR}=f_\text{HR}=1,~f_\text{incon}=10$}. 
\section{Experiments}\label{sec:experiments}
\begin{figure*}[ht!]
\centering
\includegraphics[width=\textwidth]{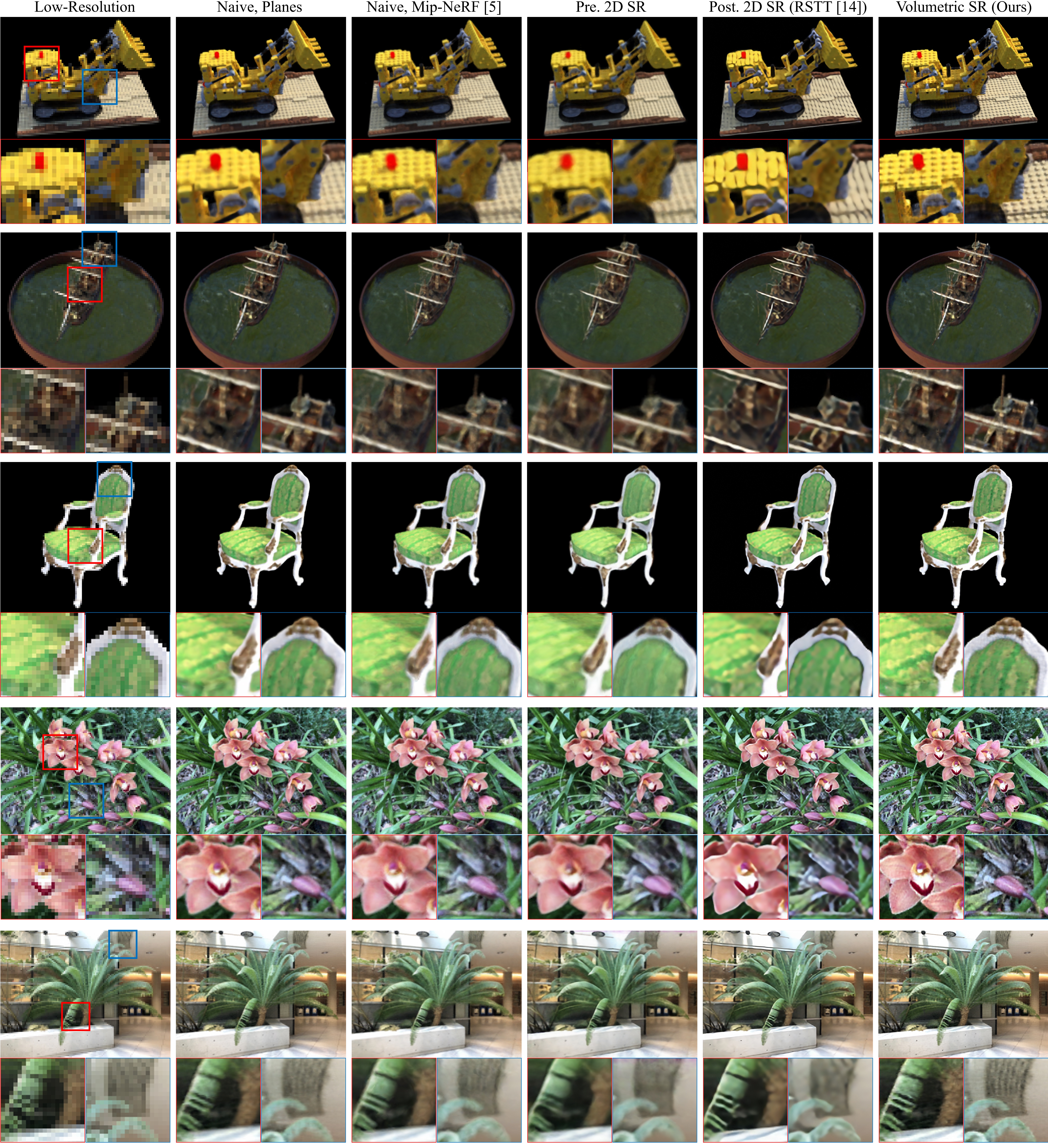}
\caption{\label{fig:results}\textbf{Qualitative super-resolution evaluation.}
We visually compare the performance of our method (right) to performance by leading SR methods and baselines, when super resolving novel low-resolution views (left). Our Neural Volume SR approach is able to reconstruct finer details and smaller structures compared to all other methods. 
While merely interpolating the feature planes in the volumetric domain (2\textsuperscript{nd} from right) already yields a quality gain,
applying our SR model $F$ on the planes achieves a significant improvement.
}
\end{figure*}

We next demonstrate the advantage of using our volumetric approach for super-resolving synthetic, as well as real 3D scenes captured in LR.
To this end, we use the sets of eight synthetic scenes and eight real scenes available on the NeRF~\cite{mildenhall2020nerf} project page, as well as additional $21$ scenes rendered from 3D Blender models available online\footnote{Links to the these freely available Blender models are provided in the Supp. material. Our entire project code will become available online.}.
These scenes depict complex objects with fine details and textures, which are crucial for demonstrating and evaluating the SR quality.
For all synthetic scenes, we use the $100$ available training images with dimensions $400\times400$ as the ground truth HR captures ${\cal I}_{\text{hr}}^s$, and then downsample each image by a factor of $4$ using the default bicubic kernel to simulate the set of LR captures corresponding to the scene, ${\cal I}_{\text{lr}}^s$.
Each of our real world scenes has a couple of dozen available images with varying dimensions, which we resize to obtain similar dimensions as in the synthetic case.

For quantitatively evaluating on synthetic scenes, we train our model (including both representation model $G$ and SR model $F$) on $25$ scenes, while holding out the remaining four (`mic', `ship', 'chair' and `lego') scenes for evaluation.
We evaluate all methods on $200$ test images corresponding to novel views of each of these unseen scenes, while comparing our approach with methods and baselines representing several different approaches, by using a set of evaluation metrics. 
We use the same evaluation protocol with real scenes, using six scenes for training and the remaining unseen two ('fern' and 'orchids') for evaluation.
We report the results on synthetic and real scenes separately in Tabs.~\ref{tab:reconst} and~\ref{tab:real_reconst}, respectively, and elaborate next on the metrics we employ, as well as on the baselines we compare to.

\myparagraph{Evaluation Metrics.}
We use PSNR and SSIM to measure how well methods minimize the error with respect to the corresponding GT HR images. To evaluate perceptual quality, we employ the LPIPS~\cite{zhang2018lpips} score, where lower means better. 

\myparagraph{Baselines.}
We compare our approach to several baseline alternatives. We first compare with the naive approach of using models trained on the LR image sets (which lack HR content) to render HR novel views, without attempting to reconstruct the missing HR content. We repeat this experiment twice, once using our quadri-plane model (`Planes') and a second time using the Mip-NeRF~\cite{Barron_2021mipnerf} method (`Mip-NeRF'), which was designed to handle multiple image resolutions for reducing aliasing and blur.

We additionally compare with alternative ways to reconstruct the missing HR content, by harnessing existing methods for 2D super resolution as pre-processing or post processing steps: In the first (denoted 'pre. 2D SR') we apply a 2D SR method on the LR training images and use the resulting super-resolved images to learn a Mip-NeRF model for the scene, while in the second (denoted 'post. 2D SR') we use the 2D SR method to super-resolve LR images rendered by a Mip-NeRF model which was learned from the LR images.
For the 2D SR step we use 
the state-of-the-art SwinIR method~\cite{liang2021swinir} which employs visual transformers. In the post-SR case, we also compare to using the prominent SRGAN~\cite{ledig2017srgan} method\footnote{We use the distortion-minimizing training configuration (without an adversarial loss) from SRGAN to train their SRResNet architecture, to allow a fair comparison with our method, which is trained to minimize distortion with respect to the GT images.} or the EDSR method~\cite{lim2017edsr}, whose network architecture we employ for our model $F$. 
Finally, since one could view the LR scene captures as pertaining to an LR video sequence, we further experiment with post-SR using a state-of-the-art method for video spatio-temporal SR~\cite{geng2022video_sr}. For the comparison to be unbiased, we train all 2D SR models from scratch using the same images from ${\cal S}_t$ that we use for training our method.

\begin{table}[t!]
    \centering
    \resizebox{1.03\columnwidth}{!}{
    \begin{tabular}{clccc}
        \toprule
        Approach&Method&PSNR (dB)$(\uparrow)$&SSIM $(\uparrow)$& LPIPS $(\downarrow)$\\
        \midrule
        \multirow{2}{*}{Naive} 
        & Mip-NeRF~\cite{Barron_2021mipnerf} & $27.5$ & $0.892$&$0.165$\\
        & Planes & $27.4$ & $0.892$& $0.153$ \\
        \midrule
        pre. 2D SR& SwinIR~\cite{liang2021swinir} & $25.6$  & $0.872$&$0.183$ \\
        \midrule 
        \multirow{4}{*}{post. 2D SR} & EDSR~\cite{lim2017edsr} & $25.9$  & $0.876$& $0.148$\\
        & SRGAN~\cite{ledig2017srgan} & $24.5$  & $0.863$&$0.149$ \\
        & SWinIR~\cite{liang2021swinir} & $25.1$  & $0.863$&$0.183$\\
        & RSTT~\cite{geng2022video_sr} &  $26.3$  & $0.877$&$0.139$\\
        \midrule
        Volume. SR & Ours & $\textbf{28.5}$ & $\textbf{0.911}$& $\textbf{0.099}$ \\
        \bottomrule
    \end{tabular}
    }
    \vspace{0pt}
    \caption{\textbf{Evaluation on synthetic scenes.} We evaluate all methods on 800 test images corresponding to novel views from 4 scenes unseen during training. 
    The reported SSIM and PSNR (higher is better) and LPIPS (lower is better) values reveal a significant advantage of the proposed volumetric SR approach over all baseline approaches.
    Please refer to the text for details.
    }
    \label{tab:reconst}
\end{table}

\myparagraph{Results.}
The quantitative comparisons reported in Tabs.~\ref{tab:reconst} and~\ref{tab:real_reconst} indicate the significant advantage of the proposed approach in terms of reconstruction quality (PSNR, SSIM) and perceptual quality (LPIPS) over all baselines, when evaluated on synthetic and real world scenes alike.

An additional inherent advantage of our approach, which super-resolves the volumetric representation itself, is its guaranteed across-view consistency. This advantage does not reflect in the quantitative results brought here, nor can it be conveyed in the still images depicted in Fig.~\ref{fig:results}. We therefore kindly refer readers to the videos provided in the Supp., which demonstrate this advantage over other baselines that perform SR in post-processing.

Interestingly, the results suggest that utilizing 2D SR methods (trained using the same data) actually has a negative effect on reconstruction performance, even compared to the naive baselines. We attribute this to the significant inconsistency resulting from independently processing different scene views.

Finally, we demonstrate the advantage of our approach visually in Fig.~\ref{fig:results}, where we compare novel test scene views rendered by our approach vs. the top performing baselines.
\myparagraph{Ablations and Limitations.}
We further examine the effect of different components and hyper-parameter choices in our framework in a series of ablation experiments. We find that the resolution of the positional feature planes should be set to $1$-$2$ times the resolution of the captured images set, while diverging from this range by a factor of $4$ (in either direction) yields a drop of $\sim1.7$ dB in PSNR, as we demonstrate visually in Fig.~\ref{fig:plane_res}. Similarly, we analyse the significance of the 4\textsuperscript{th} (view-direction) plane $P_\text{dir}$ introduced in this work, and find that it is responsible to an increase of almost $3$ dB in PSNR, which aligns with the visual demonstration in Fig.~\ref{fig:viewdir_effect}. Please refer to the Supp. for detailed results and additional ablation experiments.
\begin{table}[t!]
    \centering
    \resizebox{1.03\columnwidth}{!}{
    \begin{tabular}{clccc}
        \toprule
        Approach&Method&PSNR (dB)$(\uparrow)$&SSIM $(\uparrow)$& LPIPS $(\downarrow)$\\
        \midrule
        \multirow{2}{*}{Naive} 
        & Mip-NeRF~\cite{Barron_2021mipnerf} & $21.9$ & $0.657$&$0.401$\\
        & Planes & $22.8$ & $0.714$& $0.373$ \\
        \midrule
        pre. 2D SR& SwinIR~\cite{liang2021swinir} & $21.7$  & $0.659$&$0.430$ \\
        \midrule 
        \multirow{4}{*}{post. 2D SR} & EDSR~\cite{lim2017edsr} & $21.8$  & $0.664$& $0.389$ \\
        & SRGAN~\cite{ledig2017srgan} & $21.7$  & $0.657$&$0.386$ \\
        & SWinIR~\cite{liang2021swinir} & $21.6$  & $0.650$&$0.437$\\
        & RSTT~\cite{geng2022video_sr} &  $22.0$  & $0.679$&$0.377$\\
        \midrule
        Volume. SR & Ours & $\textbf{23.4}$ & $\textbf{0.747}$& $\textbf{0.255}$ \\
        \bottomrule
    \end{tabular}
    }
    \vspace{0pt}
    \caption{\textbf{Evaluation on real-world scenes.} All methods are evaluated on all available $45$ images corresponding to two real world scenes. For obtaining the `Naive: Planes' and `Ours' results, we use our model trained on views corresponding to other six scenes. Please refer to the text for details.
    }
    \label{tab:real_reconst}
\end{table}

Finally, like most super-resolution methods, ours too is susceptible to non-ideal capturing conditions. We analyze their effect on performance compared to the other baselines in the Supplementary.

\section{Conclusion}
In this work, we introduce and investigate the novel task of neural volume super-resolution -- rendering high-resolution views corresponding to a scene captured at low resolution. 
We propose a neural scene representation model that utilizes 2D feature planes, which we learn using LR scene captures. We then render novel, high-resolution views corresponding to the scene by super-resolving the 2D feature planes using a CNN model trained on a dataset of 3D scenes. 
We validate our approach both quantitatively and qualitatively on a diverse set of unseen synthetic and real-world scenes, and show a significant increase in performance compared to various baseline approaches in terms of view consistency, as well as in terms of reconstruction and perceptual quality, which demonstrate the major advantages of operating in the volumetric representation domain.
Beyond the advantage in performing volumetric scene super-resolution, our work reveals the potential of this approach for reconstructing scenes captured under other types of realistic degradations (\eg noise, blur), which we hope to investigate in future work.

\ifarXiv
\newpage
\noindent\textbf{Acknowledgment}

\noindent This project has received funding from the European Union’s Horizon 2020 research and innovation programme under the Marie Skłodowska-Curie grant agreement No 945422.
\fi
{\small
\bibliographystyle{ieee_fullname}
\bibliography{egbib}
}
\ifarXiv
\onecolumn
\appendixpage
\appendix
\section{Additional Evaluations}

This section provides additional qualitative and quantitative evaluations that evaluate the proposed method in comparison to existing methods for volumetric super-resolution.
\subsection{Video Comparisons for Evaluating Across View-direction Consistency}
The results in Tab.~\ref{tab:reconst} demonstrate the advantage of our approach in terms of reconstruction quality, as measured independently on each rendered novel view. 
However, a key advantage of directly super-resolving the volumetric representation of the scene is that the added high-frequency, fine details are consistent across different views. 
While this consistency is hard to assess when examining each output independently, it becomes apparent when merging the outputs into a video trajectory, where geometric inconsistencies across adjacent viewing direction manifest themselves as prominent fidgeting-like temporal artifacts.
Therefore, to demonstrate the advantage of our volumetric SR approach, we compile a series of `fly-over' videos corresponding to the rendered scenes. 
Please refer to the attached video files `RealWorldScenes.mp4' and `SyntheticScenes.mp4' to view side-by-side comparisons of results rendered through different methods operating in 3D (including ours) with results by methods that apply 2D SR (including video SR~\cite{geng2022video_sr}) in post processing. Please note the sharper results produced by our method, as well as the prominent jittering artifacts introduced by 2D SR post processing baselines.

\subsection{Effect of Non-Ideal Capturing Conditions}
Like most super-resolution methods, ours too is susceptible to degradations such as noise or non-ideal (typically unknown) downsampling kernels (PSFs). We perform experiments to quantify the effect of these conditions on the performance of our method, compared with the performance of the top performing baseline methods from Tab.~\ref{tab:reconst} under the same settings.

To this end, we simulate capturing with noise (typically under low-lighting conditions) by adding random Gaussian noise with $\sigma=5$ to the set of LR image captures ${\cal I}_\text{lr}^s$ corresponding to a test scene $s$. Similarly, we simulate the non-ideal PSF case in a separate experiment by using a Gaussian blur kernel with a standard deviation of $\sigma=1$ pixel for downsampling the GT HR images in ${\cal I}_\text{hr}^s$ when creating the corresponding set of LR inputs ${\cal I}_\text{lr}^s$.

We report the results from the unknown PSF and additive noise experiments to the left and right of the slash sign in Tab.~\ref{tab:noise_reconstruct}, respectively. Compared to the ideal setting case, the performance of all three compared baselines is reduced considerably, bringing them to perform at par under these conditions.
In the case of our method, we conjecture that the reduced performance can mainly be attributed to the undermining effect that the non-ideal capturing may have over the minimization of the ${\cal L}_\text{incon}$ loss term. We believe addressing these interesting challenges can be a topic for future work.

\begin{table}[ht!]
    \centering
    \begin{tabular}{clccc}
        \toprule
        Approach&Method&PSNR (dB)$(\uparrow)$&SSIM $(\uparrow)$& LPIPS $(\downarrow)$\\
        \midrule
        \multirow{2}{*}{Naive} 
        & Mip-NeRF~\cite{Barron_2021mipnerf} & $\textbf{26.68}/\textbf{26.87}$ & $\underline{0.872}/\underline{0.762}$&$0.205/\textbf{0.190}$\\
        & Planes & $26.27/26.25$ & $0.869/\textbf{0.800}$& $\underline{0.201}/0.260$ \\
        \midrule
        Volume. SR & Ours & $\underline{26.64}/\underline{26.60}$ & $\textbf{0.875}/0.726$& $\textbf{0.189}/\underline{0.249}$ \\
        \bottomrule
    \end{tabular}
    \caption{\textbf{Non-ideal capturing conditions for ${\cal I}_\text{lr}^s$.} Performance of our method compared with the leading baselines from Tab.~\ref{tab:reconst} when simulating unknown, non-ideal downsampling kernel (left values) or additive noise (right values). We use a Gaussian blur downsampling kernel (with $\sigma=1$ pixel) and an additive Gaussian noise (with $\sigma=5$ gray levels) for creating the LR image sets in these simulations, respectively. Experiment settings are identical to those in Tab.~\ref{tab:reconst}. Please refer to the text for more details.
    }
    \label{tab:noise_reconstruct}
\end{table}

\section{Ablation Experiments}
To understand the significance of different elements and design choices in our framework, we perform a series of ablation studies. We next elaborate on the settings, report the results and derive conclusion from each of these experiments.

\subsection{Significance of View-direction (4\textsuperscript{th}) Plane}
As we describe in Sec.~\ref{sec:planar_representation}, Chan \etal~\cite{chan2022eg3d} use a tri-plane model constituting three positional feature planes $\{P_{xy},P_{xz},P_{yz}\}$ for representing human face scenes.
We propose to add a 4\textsuperscript{th} feature plane $P_\text{dir}$ whose two axes span the 2D space of possible viewing directions for each point $p$. We do that in order to accommodate effects like non-uniform scene illumination, and subsequently to improve reconstruction performance, as we show in Fig.~\ref{fig:viewdir_effect}.

To further evaluate the effectiveness of $P_\text{dir}$, we compare the proposed quadri-plane model with a tri-plane model comprising only the three positional planes. Note that we did not include SR model $F$ in this ablation experiment. We present a qualitative side-by-side comparison in the attached `EffectOfViewdirPlane.mp4' video file, and provide a quantitative performance evaluation in Tab.~\ref{tab:viewdir_plane}, where we measure the performance of the two models on the task of reconstructing $1400$ novel views corresponding to $7$ scenes, using the PSNR, SSIM and LPIPS metrics. The results by all metrics, including an almost $3dB$ increase in PSNR, indicate a clear advantage for the proposed quadri-plane model.
\begin{table}[ht]
    \centering
    \begin{tabular}{lcccc}
        \toprule
        $P_\text{dir}$& PSNR $(db)(\uparrow)$&SSIM $(\uparrow)$& LPIPS $(\downarrow)$\\
        \midrule 
        With &  $\textbf{28.7}$ &$\textbf{0.898}$ & $\textbf{0.078}$\\
        Without & $25.8$ & $0.867$ & $0.122$\\
        \bottomrule
    \end{tabular}
    \caption{\textbf{Effect of 4\textsuperscript{th} feature plane $P_\text{dir}$.}
    Reconstruction performance of our representation model $G$ with vs. without utilizing the 4\textsuperscript{th} view-direction plane $P_\text{dir}$. Results indicate a significant advantage for our quadri-plane (top row) as features in $P_\text{dir}$ are able to account for view-dependent effects such as non-uniform illumination and reflections.
    }
    \label{tab:viewdir_plane}
\end{table}

\subsection{Effect of Feature Planes Resolution $N$}
The spatial dimensions $N$ of the positional feature planes in the proposed quadri-plane representation model strongly affects representation performance, as we discuss in Sec.~\ref{sec:planar_representation} and demonstrate in Fig.~\ref{fig:plane_res}. 
Too small planes cannot represent high visual frequencies. On the other hand, using $N$ values much larger than the resolution of the captured scene images results in visual artifacts when rendering novel views, as some plane regions remain uncovered by the information in ${\cal I}^s$.
As a rule of thumb, we set $N$ to be one to two times the resolution of the captured images $I\in{\cal I}^s$. To validate this choice, we compare reconstruction performance over $1400$ novel views corresponding to $7$ scenes when setting $N$ to be a quarter, the same as or $4$ times the resolution of $I$. The representations in this experiment were learned using images $I\in{\cal I}^s$ of size $400\times400$, and we use each learned feature plane as is, without super-resolving it with $F$. The results in Tab.~\ref{tab:plane_res} suggest setting an appropriate resolution value $N$ has a significant effect, and support our rule of thumb.
\begin{table}[!ht]
    \centering
    \begin{tabular}{ccccc}
        \toprule
        $N/$resolution($I\in{\cal I}^s$) & PSNR $(db)(\uparrow)$&SSIM $(\uparrow)$& LPIPS $(\downarrow)$\\
        \midrule 
        $1/4$ & $27.3$ & $0.894$ & $0.126$\\
        $1$ &  $\textbf{28.8}$ &$\textbf{0.920}$ & $\textbf{0.075}$\\
        $4$ & $27.2$ & $0.879$ & $0.127$\\
        \bottomrule
    \end{tabular}
    \caption{\textbf{Effect of positional feature planes resolution $N$.}
    Performance of our representation model when using positional feature planes of different sizes $N$. Results suggest this parameter strongly effect performance, as supported by the qualitative comparison in Fig.~\ref{fig:plane_res}.
    }
    \label{tab:plane_res}
\end{table}

\subsection{Effects of Inconsistency Loss Term ${\cal L}_\text{incon}$ and SR Model $F$}
We conduct additional ablation studies to analyse the role of these central components in the proposed volumetric SR approach. To this end, we
repeat the evaluation experiment reported in Tab.~\ref{tab:reconst} twice more: Once without performing the second training phase described in Sec.~\ref{sec:implementation} (denoted `No ${\cal L}_\text{incon}$' in Tab.~\ref{tab:ablation_recon}), and a second time where we discard the feature-planes SR model $F$ (denoted by `No SR'). In the latter experiment we perform the second training phase while optimizing only the weights of the decoder pair $D$, which take as input feature vectors $v_p$ extracted directly from the learned low-resolution feature planes.

We report the results in Tab.~\ref{tab:ablation_recon}, where we also bring as reference the results by the Mip-NeRF~\cite{Barron_2021mipnerf} baseline, by the naive baseline utilizing our quadri-plane model (denoted `LR feature planes (Naive)') and the results by our full proposed configuration (bottom row). 
Note the significant drop in performance when eliminating either the SR model $F$ ($-0.7dB$ in PSNR) or the practice of minimizing ${\cal L}_\text{incon}$ ($-1.4dB$ in PSNR), indicating the crucial role played by both of them.
\begin{table}[ht!]
    \centering
    \begin{tabular}{clccc}
        \toprule
        Approach&Method&PSNR (dB)$(\uparrow)$&SSIM $(\uparrow)$& LPIPS $(\downarrow)$\\
        \midrule
        NeRF& Mip-NeRF~\cite{Barron_2021mipnerf} & $27.5$ & $0.892$&$0.165$\\
        \midrule
        \multirow{4}{*}{\makecell[c]{Volumetric \\SR \\ (Ours)}} 
        & No SR & $27.8$ & $0.899$& $0.126$ \\
        & No ${\cal L}_\text{incon}$ & $27.1$ & $0.887$& $0.160$ \\
        & LR Feature Planes (Naive) & $27.4$ & $0.892$& $0.153$ \\
        & Proposed Configuration & $\textbf{28.5}$ & $\textbf{0.911}$& $\textbf{0.099}$ \\
        \bottomrule
    \end{tabular}
    \vspace{0pt}
    \caption{\textbf{Ablation experiments.} Comparing performance by the proposed method (bottom row) with its ablated variants when either excluding feature planes SR model $F$ (denoted `No SR'), or not minimizing ${\cal L}_\text{incon}$. Experiment settings are identical to those in Tab.~\ref{tab:reconst}. The significant performance drop indicate the importance of both of components to the performance of our method.
    Please refer to the text for more details.
    }
    \label{tab:ablation_recon}
\end{table}

\section{Quantifying Across View-direction Consistency}\label{sec:AVI}
\begin{table}[t!]
    \centering
    \begin{tabular}{clc}
        \toprule
        Approach&Method&AVI$(\downarrow)$\\
        \midrule
        \multirow{2}{*}{Naive} 
        & Mip-NeRF~\cite{Barron_2021mipnerf} & $\textbf{1.63}$\\
        & Planes & $1.97$\\
        \midrule
        pre. 2D SR& SwinIR~\cite{liang2021swinir} & $1.71$\\
        \midrule 
        \multirow{4}{*}{post. 2D SR} & EDSR~\cite{lim2017edsr} & $2.82$\\
        & SRGAN~\cite{ledig2017srgan}  & $2.92$\\
        & SWinIR~\cite{liang2021swinir} &  $2.61$\\
        & RSTT~\cite{geng2022video_sr} &$2.51$\\
        \midrule
        Volume. SR & Ours & $1.86$\\
        \bottomrule
    \end{tabular}
    \vspace{0pt}
    \caption{\textbf{Across View-direction Inconsistency (AVI).} Inconsistency scores (lower is better) are evaluated across $800$ test frames corresponding to four synthetic scenes, following the experiment configuration from Tab.~\ref{tab:reconst}. 
    The results reveal the inherent advantage of methods operating in the 3D domain (including ours) over methods performing 2D SR in post processing.
    Please refer to the text for more details.
    }
    \label{tab:avi}
\end{table}

To allow quantitatively evaluating this advantage of our approach, we next introduce a new Across View-direction Inconsistency (AVI) metric which aims to quantify the degree of geometrical inconsistency between every pair of test frames corresponding to adjacent viewing directions.
To this end, we harness the GT optical flow that can be extracted for every adjacent frames pair in our synthetically rendered scenes.

We bring the formula for computing the AVI score and describe in detail each of its parts in Sec.~\ref{sec:avi_detail}. We report AVI scores corresponding to our approach, as well as to all other compared baselines, in Tab.~\ref{tab:avi}.
The results reveal 
the inherent advantage of methods operating in the 3D domain. 
In terms of consistency, our method scores in between those baselines and significantly better than all methods performing 2D SR in post processing.

\subsection{Calculating the AVI Score}\label{sec:avi_detail}
\yb{I should fix the PatchNorm operator $\textbf{g}$ definition to reflect the fact that it operates on both images jointly, as it filters out patch pairs when the STD of either one of them falls below $\epsilon$.}
We use the following formula to calculate the AVI score corresponding to a scene $s$:
\begin{equation}\label{eq:inconsistency}
    \text{AVI} = \frac{1}{n}\sum_{k=0}^{\|{\hat{\cal I}}_{\text{sr}}\|-1}M_{k,k+1}\cdot \|\textbf{g}\Bigl(\text{Warp}\bigl({\hat i}_{\text{sr}}^k\bigr)\Bigr)-\textbf{g}\Bigl({\hat i}_{\text{sr}}^{k+1}\Bigr)\|_2.
\end{equation}

The AVI for each scene is computed over the set $\hat{\cal I}_{\text{sr}}$ of rendered HR views ${\hat i}_{\text{sr}}^k$ corresponding to all test frames available for this scene. Since we have access to the Blender models used to generate the synthetic scene datasets, we are able to compute ground truth optical flow motion vectors both for the forward flow $f_{k\rightarrow k+1}$ and for the backward flow $f_{k+1 \rightarrow k}$ between all adjacent view-direction pairs $k,k+1$. We utilize the \emph{backward flow} to warp each image to the next adjacent viewing direction via
\begin{equation}
    \text{Warp}(\hat{i}_{\text{sr}}^k)(r) = \hat{i}_{\text{sr}}^k(r + f_{k+1 \rightarrow k}(r)),
\end{equation}
where $r$ denotes the 2D pixel location. 

Ideally we would want the local differences between each frame and its warped adjacent view direction counterpart to be independent of whether a scene is smooth or highly textured. To this end, we apply a pre-processing operator $\textbf{g}$ on each of the two frames independently, before calculating the difference between them:
\begin{equation}\label{eq:process4avi}
    \textbf{g}(i)=\text{PatchNorm}\bigl(\text{Im2patch}(i)\bigr).
\end{equation}
Here $\text{Im2patch}(\cdot)$ extracts all overlapping $7\times7$ patches from an RGB image and $\text{PatchNorm}(\cdot)$ acts by discarding very smooth patches with standard deviation below a minimum value $\epsilon=3/255$ (which are typically governed by noise) and normalizing the standard deviation of all other patches:
\begin{equation}\label{eq:patch_norm}
    \text{PatchNorm}(p)=
    \begin{cases}
     0, &\text{if std}(p)<\epsilon\\
     p/\text{std}(p), &\text{otherwise}
    \end{cases}.
\end{equation}

Even though we have access to the ground truth motion vectors, the warping can still introduce ghosting artifacts due to occlusions, i.e., pixels in image $k$ that have no correspondence in image $k+1$. To prevent errors stemming from the ghosting artifacts from dominating the AVI metric, we introduce an error mask $M_{k, k+1}$, computed as follows: First, by using the \emph{forward flow}, we count how many pixels get mapped to the same pixel $r$:
\begin{equation}
    \text{Count}_{k, k+1}(r) = \#\{ q | \lfloor q + f_{k\rightarrow k+1}(q)\rfloor = r\}.
\end{equation}
Pixels from viewing direction $k+1$ for which this count is zero do not have a correspondence in viewing direction $k$. Thus, our mask is defined as
\begin{equation}
    \Tilde{M}_{k, k+1}(r) = \begin{cases}
     0, &\text{if Count}_{k, k+1}(r) = 0\\
     1, &\text{otherwise}
    \end{cases}.
\end{equation}
Since the floor operation in the count computation can still lead to some ghosting due to subpixel occlusions, we apply a morphological closing followed by erosion to the mask, using a "cross" structuring element of radius 1. The final mask in Eq.~\ref{eq:inconsistency} is thus given by
\begin{equation}
    M_{k, k+1} = \text{erosion}(\text{closing}(\Tilde{M}_{k, k+1})).
\end{equation}
When computing the AVI score, we only consider unmasked elements for the mean computation, i.e. $n = \#\{ r | M_{k, k+1}(r) = 1\}$ in Eq.~\ref{eq:inconsistency}.
The effects of the raw mask and the different morphological operators are visualized in Fig.~\ref{fig:masks}.
Note that we use the same motion vectors, and thus also the same masks, when computing the AVI score for each of the compared methods.
\begin{figure*}
\centering
\begin{tabular}{c c c c}
    \includegraphics[width=0.22\textwidth]{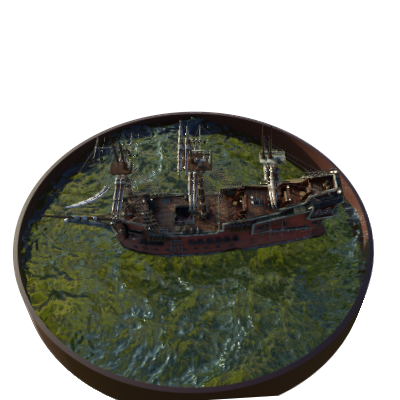}&
    \includegraphics[width=0.22\textwidth]{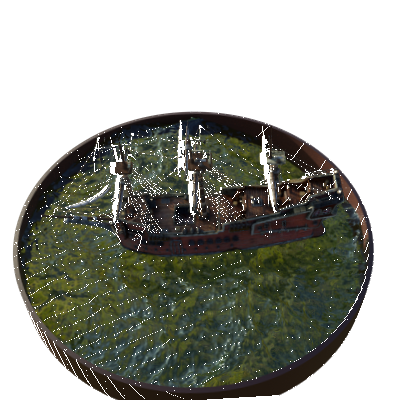}&
    \includegraphics[width=0.22\textwidth]{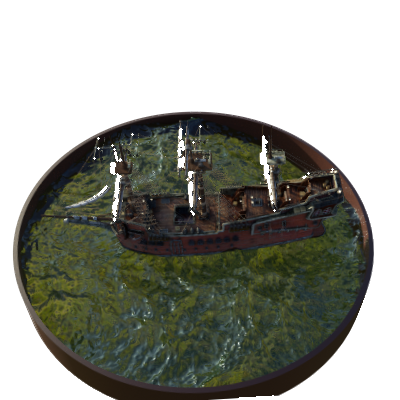}&
    \includegraphics[width=0.22\textwidth]{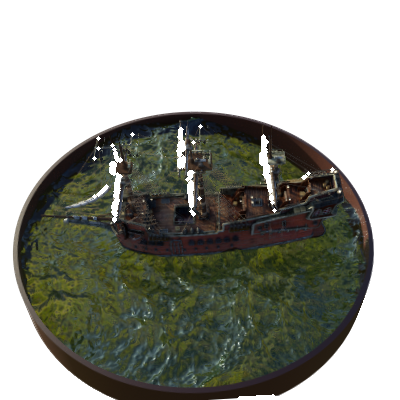}\\
    (a) Without masking &
    (b) With $\Tilde{M}_{k, k+1}$&
    (c) With closing &
    (d) With closing + dilation
\end{tabular}
\caption{\textbf{Effect of Mask $M_{k, k+1}$.} We use masking to  remove ghosting artifacts (\eg the ship's masts) in the warped image (a) prior to calculating the error. Here we illustrate the effect of using the mask with or without processing it with morphological image operators (b-d), and show that applying the fully processed mask yields significantly reduced ghosting artifacts (d).}
\label{fig:masks}
\end{figure*}
\section{Synthetic Dataset 3D Scene Models}
Our approach learns to super-resolve the volumetric representations by training on a set of training scenes ${\cal S}_t$. Being a deep-learning-based approach, it requires sufficient amount of training data to be able to generalize from the training scenes to unseen scenes. To meet this data requirement, 
we augment the synthetic scenes from \cite{mildenhall2020nerf} with 8 additional sequences generated from the following freely available Blender scenes
\begin{itemize}
    \item \url{https://free3d.com/3d-model/bugatti-chiron-2017-model-31847.html}
    \item \url{https://free3d.com/3d-model/black-dragon-rigged-and-game-ready-92023.html}
    \item \url{https://free3d.com/3d-model/professional-scene-with-coca-cola-bottle-57999.html}
    \item \url{https://free3d.com/3d-model/gibson-es-335-816888.html}
    \item \url{https://free3d.com/3d-model/fuzzy-bear--10429.html}
    \item \url{https://free3d.com/3d-model/harley-davidson-low-rider-84874.html}
    \item \url{https://free3d.com/3d-model/holiday-beach-cartoon-scene-431138.html}
    \item \url{https://free3d.com/3d-model/donut-503129.html}
    \item \url{https://www.blendswap.com/blend/30606}
    \item \url{https://www.blendswap.com/blend/30328}
    \item \url{https://www.blendswap.com/blend/30072}
    \item \url{https://www.blendswap.com/blend/29594}
    \item \url{https://www.blendswap.com/blend/29435}
    \item \url{https://www.blendswap.com/blend/29080}
    \item \url{https://www.blendswap.com/blend/29508}
    \item \url{https://www.blendswap.com/blend/29461}
    \item \url{https://www.blendswap.com/blend/17636}
    \item \url{https://www.blendswap.com/blend/16763} (we removed the tender)
    \item \url{https://www.blendswap.com/blend/16155}
    \item \url{https://www.blendswap.com/blend/13078}
    \item \url{https://www.blendswap.com/blend/5364}

\end{itemize}

Training, validation and test sequences were generated using the Blender script provided by Mildenhall \etal~\cite{mildenhall2020nerf}.
\fi

\end{document}